\documentclass[lettersize,journal]{IEEEtran}
\usepackage{amsmath,amsfonts}
\usepackage{algorithmic}
\usepackage{array}
\usepackage[caption=false,font=normalsize,labelfont=sf,textfont=sf]{subfig}
\usepackage{textcomp}
\usepackage{stfloats}
\usepackage{url}
\usepackage{verbatim}
\usepackage{graphicx}
\usepackage{cite}
\usepackage{amsmath,amssymb,amsfonts}
\usepackage{multirow}    
\usepackage{diagbox}
\usepackage{booktabs} 
\usepackage{algorithm}
\def\BibTeX{{\rm B\kern-.05em{\sc i\kern-.025em b}\kern-.08em
    T\kern-.1667em\lower.7ex\hbox{E}\kern-.125emX}}
\usepackage{balance}
\begin{document}
\title{Unsupervised Unfolded rPCA (U\textsuperscript{2}-rPCA): Deep Interpretable Clutter Filtering for Ultrasound Microvascular Imaging}

\author{Huaying Li, Chuling Ye, Manfei Liao, Xiaobo Qu, \IEEEmembership{Senior Member, IEEE}, Liansheng Wang, \IEEEmembership{Member, IEEE}, and Yinran Chen, \IEEEmembership{Member, IEEE}
\thanks{Manuscript received XXX XX, 2024. This work was supported by the National Natural Science Foundation of China (No. 62471416), the Fujian Provincial Natural Science Foundation of China (No. 2025J09001 and 2024J01003), the Natural Science Foundation of Xiamen, China (No. 3502Z202473004), and the Fundamental Research Funds for the Central Universities (No. 20720240075). (Corresponding authors: Liansheng Wang and Yinran Chen).}
\thanks{H. Li, C. Ye, and M. Liao are with the Fujian Key Laboratory of Urban Intelligent Sensing and Computing, School of Informatics, Xiamen University, Xiamen 361005, China.}
\thanks{X. Qu is with the School of Electronic Science and Engineering, Fujian Provincial Key Laboratory of Plasma and Magnetic Resonance, Xiamen University, Xiamen 361005, China.}
\thanks{L. Wang and Y. Chen are with the Department of Computer Science and Technology, School of Informatics, and the National Institute for Data Science in Health and Medicine, Xiamen University, Xiamen 361005, China (E-mail: lswang@xmu.edu.cn, yinran\_chen@xmu.edu.cn).}}


\maketitle

\begin{abstract}
High-sensitivity clutter filtering is a fundamental step in ultrasound microvascular imaging. Singular value decomposition (SVD) and robust principal component analysis (rPCA) are the main clutter filtering strategies. However, both strategies are limited in feature modeling and separation of tissue and blood flow for high-quality microvascular imaging. Recently, deep learning-based clutter filtering has shown potential in more thoroughly separating tissue and blood flow signals. However, the existing supervised filters face the lack of interpretability and the training ground truth. While the interpretability issue can be addressed by algorithm deep unfolding, the training ground truth remains unsolved. This paper proposes an unsupervised unfolded rPCA (U\textsuperscript{2}-rPCA) method that preserves mathematical interpretability and is insusceptible to learning labels. Specifically, U\textsuperscript{2}-rPCA is unfolded from an iteratively reweighted least squares (IRLS) rPCA baseline with intrinsic low-rank and sparse regularization. In addition, a sparse-enhancement unit is plugged into the network to strengthen its capability to capture the sparse micro-flow signals. U\textsuperscript{2}-rPCA is like an adaptive filter that is trained with part of the image sequence and then used for the following frames. Experimental validations on a \textit{in-silico} dataset and public \textit{in-vivo} datasets demonstrated the outperformance of U\textsuperscript{2}-rPCA when compared with the SVD filter, the rPCA baseline, and another deep learning-based filter. Particularly, the proposed method improved the contrast-to-noise ratio (CNR) of the power Doppler image by 1.91 dB to 8.48 dB compared to other methods. Furthermore, the effectiveness of the building modules of U\textsuperscript{2}-rPCA was validated through ablation studies.   

\end{abstract}


\section{Introduction}
\label{sec:introduction}

Ultrasound is a fundamental medical imaging modality in clinical routine. Blood flow imaging is one of the most representative functions of ultrasound. Since microvasculature and hemodynamics are important indicators for the diagnosis and assessment of many diseases, ultrasound microvascular imaging, including ultrasound localization microscopy (ULM), has emerged as a powerful tool for visualizing and quantifying microvasculature in various laboratory and clinical applications~\cite{8396283, Yi2022ARO}.

\begin{table*}[!hbt]
    \caption{Deep Learning-based Clutter Filtering Methods For Ultrasound Microvascular Imaging}
    \label{deep-learnig}
    \renewcommand{\arraystretch}{1.2}
    \centering
    
    \begin{tabular*}{\textwidth}{@{\extracolsep{\fill}}p{1.5cm}|p{2.5cm}p{2cm}p{2.5cm}p{4cm}}
    \hline
     Category&  Authors&  Model&  Training Approach&  Training Data \\
    \hline
    \multirow{2}{*}{Unfolded}
           & Pustovalov \textit{et al}. [26] & DRPCANet &  Supervised & \textit{In-silico}\\
           & Solomon \textit{et al}. [25] & CORONA &  Supervised & \textit{In-silico} + \textit{In-vivo}~(rPCA)\\
           
    \hline
    \multirow{6}{*}{Conventional}
         & Shih \textit{et al} [28]. &  Autoencoder&  Supervised & \textit{In-silico}\\
         &  Brown \textit{et al}. [30] &  3DCNN&  Supervised &  \textit{In-vitro}~(SVD) + \textit{In-vivo}~(SVD) \\
         &  Wang \textit{et al}. [7] &  LSTM&  Supervised &  \textit{In-vivo}~(SVD)\\
         &  Ehrenstein \textit{et al}. [8] &  RA-DR\textsuperscript{2}Net&  Supervised & \textit{In-silico} \\
         &  Minhaz \textit{et al}. [29] &  2D U-Net&  Supervised & \textit{In-vivo}~(SVD) \\
         &  Ianni \textit{et al}. [27] &  3D-Res-UNet&  Supervised & \textit{In-vivo}~(SVD) \\
    \hline
    \end{tabular*}
\end{table*}

Ultrafast beamforming, clutter filtering, and denoising technologies facilitate the development of ultrasound microvascular imaging~\cite{kou2024high, demene2015spatiotemporal, 9732976}. Clutter filtering is the key to separating microvessels and tissues from the raw ultrasonic sequence. Current studies mainly employ singular value decomposition (SVD) or robust principal component analysis (rPCA) to build the clutter filters.
SVD-based filters assume that tissue and blood flow occupy different subspaces when the raw ultrasonic sequence is stacked into a Casorati matrix. The key step is to threshold the singular values of tissue and noise and recover the subspace of blood flow~\cite{demene2015spatiotemporal}. SVD has been the most popular approach for microvascular imaging. However, implementing SVD clutter filtering requires elaborate fine-tuning of thresholds. 
On the other hand, rPCA-based filters leverage the low-rankness of tissue and the sparsity of blood flow and transfer clutter filtering into low-rank and sparse decomposition~\cite{bayat2018concurrent}. Since the original form of rPCA is NP-hard, various relaxation strategies for the low-rankness and sparsity terms have been proposed. Nevertheless, rPCA-based filters are sensitive to hyperparameters. In addition, time efficiency is limited by the iterative calculations for convergence.

SVD and rPCA-based filters employ algorithmic formulations to separate tissue and blood flow signals. However, complex acoustic clutter and noise in real imaging scenarios can violate elegant mathematical assumptions. Recently, deep learning has emerged as a data-driven tool that better captures data complexity and improves microvascular clutter filtering~\cite{wang2021preliminary, ehrenstein2021rank, tabassian2019clutter}. General deep learning-based clutter filters are constructed from building blocks such as convolutional neural networks (CNNs), which lack interpretability. Alternatively, the iterative algorithms can be unfolded into successive interpretable layers~\cite{gregor2010learning}. Deep unfolding provides a concrete connection between iterative algorithms and deep neural networks~\cite{9363511,9934915}, which has attracted enormous attention from various research fields~\cite{zhang2023total,huang2021exponential,mamistvalov2021deep,khan2022unfolding,joukovsky2023interpretable,9165946,9298950}. In deep-unfolded clutter filtering, training data become critical since the current networks are trained in a supervised manner (see Table~\ref{deep-learnig}). However, the ground truth of \textit{in-vivo} tissue and blood flow signals is technically inaccessible. As a result, either \textit{in-silico} simulations or \textit{in-vivo} datasets obtained by conventional methods have to be used for training. The simulation strategy is preferred to mimic the microvasculature characteristics, such as morphology, hemodynamics, and coupling between tissue and blood flow. However, most present simulations employ simple convolutions between random-moving scatterers and a point spread function (PSF). In contrast, the \textit{in-vivo} datasets highly rely on the effectiveness of conventional SVD or rPCA clutter filtering methods.   

Deep unfolded clutter filtering shows potential in advanced ultrasound microvascular imaging. However, the quality of training data limits the performance of supervised filters. Since rPCA is an unsupervised procedure that explores the intrinsic low rankness of tissue and sparsity of blood flow, this paper proposes an Unsupervised Unfolded rPCA (U\textsuperscript{2}-rPCA) method that can (i) better separate tissue and blood flow via interpretable deep learning and (ii) be free of training ground truth. The technical contributions are clarified as follows.

\begin{itemize}
\item We construct the U\textsuperscript{2}-rPCA clutter filter based on an Iteratively Reweighted Least Squares rPCA (IRLS-rPCA) baseline. All the terms of IRLS-rPCA are formulated with the Frobenius norm so that they are directly unfolded with intrinsic low-rank and sparse regularization and trained with the equivalent (mean square error, MSE) loss functions. As a result, the network is trained in an unsupervised manner without losing interpretability. 
\item We transfer the predefined weights and penalty coefficients of IRLS-rPCA into learnable matrices and parameters to increase U\textsuperscript{2}-rPCA's ability in learning and separating the features of tissue and blood flow. Additionally, no extra effort is needed to fine-tune the penalty coefficients, which originally played an important role in the rPCA baseline. 
\item We enhance the performance of U\textsuperscript{2}-rPCA in resolving the micro blood flow signals by plugging a sparse-enhancement unit (SEU) to the unfolded layers. The unit includes a dual-frame U-Net module that assists the network in better learning the sparse spatiotemporal coherence and improving the effectiveness of blood flow extraction.
\end{itemize}


We evaluated U\textsuperscript{2}-rPCA in a \textit{in-silico} dataset and the public \textit{in-vivo} datasets. Extensive experiments demonstrated the effectiveness and interpretability of U\textsuperscript{2}-rPCA and its outperformance compared to other clutter filtering methods.

\section{RELATED WORK}
SVD-based clutter filtering requires three main steps: decomposing the raw ultrasonic sequence, thresholding the singular values, and recovering blood flow signals from its subspace. 
Following this procedure, Demené \textit{et al}.~\cite{demene2015spatiotemporal} introduced SVD to ultrafast power Doppler imaging (uPDI). 
For the two-cutoff thresholding, Song \textit{et al}.~\cite{song2016ultrasound} estimated the cutoffs using local data statistics of singular values and vectors. Baranger \textit{et al}.~\cite{baranger2018adaptive,baranger2023fast} recommended a cutoff estimator based on the spatial similarity matrix. However, the two-cutoff assumption is not rigorous since the subspaces of tissue, blood flow, and noise are not independent of each other. Each decomposed dyad may contain the contributions from different components. Waraich \textit{et al}.~\cite{waraich2019auto} employed K-means to cluster the singular ranks of blood flow for clutter filtering. We previously proposed to use competitive swarm optimization (CSO) to identify the proportion of blood flow in each singular value for improved uPDI~\cite{chen2024competitive,fang2023competitive}.


rPCA-based clutter filtering decomposes the low-rank tissue and sparse blood flow via iterative calculations until convergence. Bayat and Fatemi \textit{et al}.~\cite{bayat2018concurrent} used nuclear norm and frequency-domain $l_1$-norm to model tissue signals and blood flow signals, respectively.
Xu \textit{et al}.~\cite{xu2021robust} used a conventional $l_1$-norm for the sparse blood flow signals. This group also adopted the Cauchy norm to replace the nuclear norm and the $l_1$-norm for tissue and blood flow signals~\cite{sui2022randomized}.  
%
%
Solomon \textit{et al}.~\cite{solomon2019deep} proposed the mixed $l_{1,2}$-norm to strengthen the sparsity of blood flow signals. They also used the rPCA filtering results to supervise the training of a corresponding unfolding network.

Table~\ref{deep-learnig} illustrates that deep learning-based clutter filtering can be categorized into conventional and unfolding-based models. 
%
%
%
CORONA~\cite{solomon2019deep} is a representative of deep unfolded clutter filtering methods. It unfolds an rPCA model including a nuclear norm and a mixed $l_{1,2}$-norm into multiple layers and trains the network with \textit{in-silico} simulations and \textit{in-vivo} data obtained from the rPCA baseline. 
Pustovalov \textit{et al}.~\cite{pustovalov2022deep} unfolded a deconvolutional rPCA model into a DRPCANet to improve spatial resolution in the microvascular flow estimation.
Ianni \textit{et al}.~\cite{di2022deep} proposed a 3D-Res-UNet to learn the power Doppler reconstruction function from sparse sequences in functional ultrasound imaging.
Other classical deep neural modules, such as autoencoder~\cite{shih2019power}, 2D U-Net~\cite{minhaz2022end}, 3D CNN~\cite{brown2020deep}, LSTM~\cite{wang2021preliminary}, and deep residual reconstruction~\cite{ehrenstein2021rank} were also used for microvascular clutter filtering.
%
%
Since the abovementioned methods were trained in a supervised manner, training data significantly affect the performance of the filters. However, the \textit{in-vitro} and \textit{in-vivo} ground truth is theoretically and technically unavailable. 

\section{METHODS}
\begin{figure*}[!htb]
\centering
    \includegraphics[scale=1]{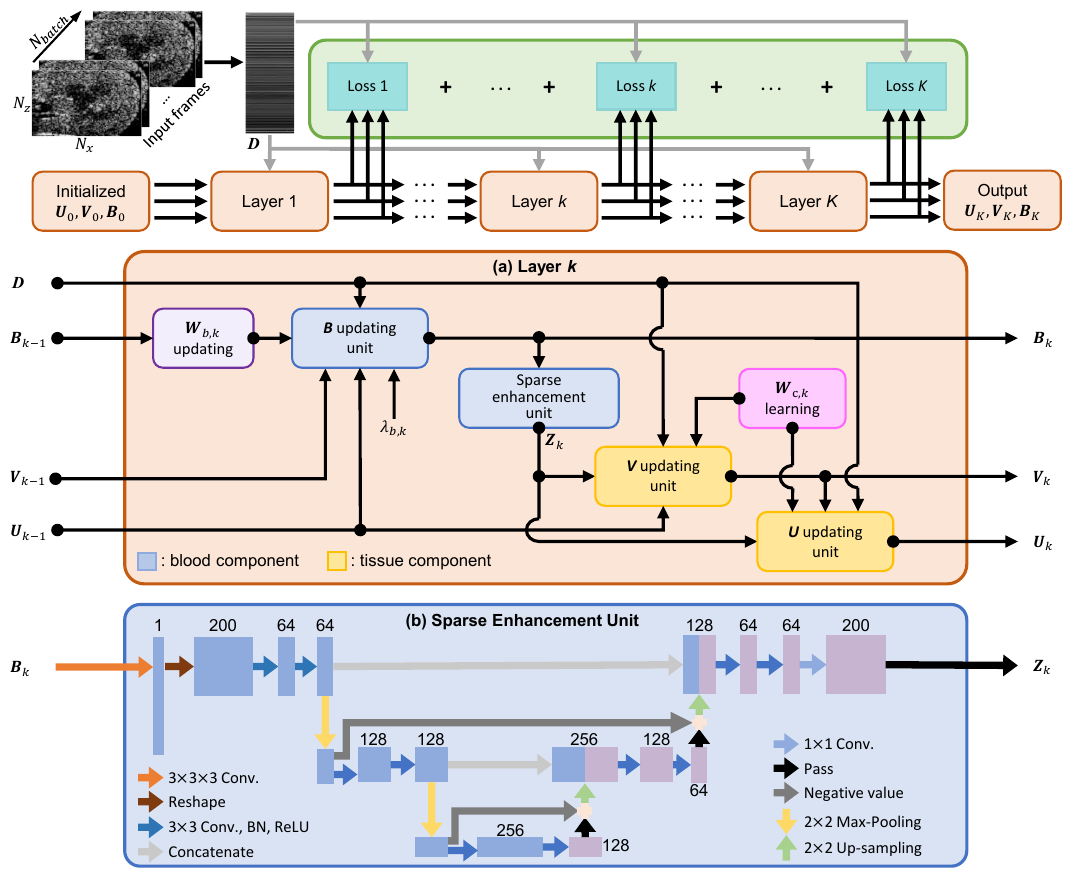}
    \caption{The framework of the U\textsuperscript{2}-rPCA clutter filter. (a) is the realization in the $k$-th layer. (b) is the architecture of the sparse-enhancement unit (SEU).}
    \label{fig:model}
\end{figure*}

\subsection{Problem Formulation}
In ultrasound microvascular imaging, a raw sequence $\boldsymbol{D}\in \mathbb{C}^{N_z, N_x, N_t}$ is acquired from ultrafast plane wave beamforming. Here, $N_x$ and $N_z$ are the number of pixels in the lateral and axial dimensions, respectively. $N_t$ is the number of frames. For clutter filtering, the raw sequence $\boldsymbol{D}$ is first reshaped into a Casorati matrix $\boldsymbol{D}\in \mathbb{C}^{N_s, N_t}$ ($N_s=N_x\times N_z$), where each column of $\boldsymbol{D}$ corresponds to a vectorized frame. The fundamental assumption is straightforward: the raw sequence is the summation of tissue signals $\boldsymbol{C}$, blood flow signals $\boldsymbol{B}$, and additive noise $\boldsymbol{N}$:
 \begin{equation}
    \boldsymbol{D}=\boldsymbol{C}+\boldsymbol{B}+\boldsymbol{N}\label{rpca}
\end{equation}
%
In the Casorati form, tissue signals $\boldsymbol{C}$ are assumed to be low-rank because they are static or moving slowly in contrast to the embedded microvessels. The blood flow signals $\boldsymbol{B}$ are dynamic within the microvascular networks, which are assumed to be spatially sparse. In addition, the spatial distribution of the microvessels is stable during the ultrafast and short acquisition. Therefore, the classical rPCA-based clutter filtering aims at solving the following low-rank and sparse decomposition:
\begin{equation}
    \min_{\boldsymbol{C},\boldsymbol{B}} \, \mathrm{rank}(\textbf{C})+\lambda  {\lVert \boldsymbol{B} \rVert}_{0} \quad
      \text{s.t.} \, \boldsymbol{C} + \boldsymbol{B} = \boldsymbol{D} \label{min_que}
\end{equation}

\noindent where $\mathrm{rank}(\cdot)$ computes the rank of a given matrix. ${\lVert \cdot \rVert}_{0}$ denotes the number of non-zero elements in a given vector or matrix. Since solving Eq.~(\ref{min_que}) is NP-hard, several strategies have been proposed to relax $\mathrm{rank}(\boldsymbol{C})$ and ${\lVert \boldsymbol{B} \rVert}_{0}$ into solvable terms~\cite{vaswani2018robust,bouwmans2018applications}, among which the nuclear norm for tissue and the $l_1$-norm for blood flow are commonly used:
\begin{equation}
    \min_{\boldsymbol{C},\boldsymbol{B}}\,  {\lVert \boldsymbol{C} \rVert}_{*}+\lambda  {\lVert \boldsymbol{B} \rVert}_{1} \quad 
    \text{s.t.} \, \boldsymbol{C} + \boldsymbol{B} = \boldsymbol{D}
    \label{nuclear_and_l1}
\end{equation}

\noindent where ${\lVert \cdot \rVert}_{*}$ computes the summation of the singular values of a given matrix, and ${\lVert \cdot \rVert}_{1}$ computes the absolute sum of all the elements of a given vector or matrix. According to~\cite{boyd2011distributed,candes2011robust}, the low-rank term is iteratively optimized by singular value thresholding, whereas the sparse term is updated by soft thresholding. However, singular value thresholding is time-consuming because SVD is needed for each iteration. More importantly, directly setting the nuclear norm and $l_1$-norm as the loss function for U\textsuperscript{2}-rPCA results in training difficulties because neither of them is differential everywhere.  

%
%
\subsection{Iteratively Reweighted Least Squares (IRLS) rPCA}
\begin{algorithm}[t]
    \caption{IRLS-rPCA Clutter Filtering}
    \label{alg:IRLS}
    \renewcommand{\algorithmicrequire}{\textbf{Input:}}
    \renewcommand{\algorithmicensure}{\textbf{Output:}}
    \begin{algorithmic}
        \REQUIRE $\boldsymbol{D}$, $d$, $\lambda_{c}$, $\lambda_{b}$, and the tolerance $\epsilon_0$ 
        \ENSURE blood flow signals $\boldsymbol{B}$    
        
        \STATE \textbf{Initialization}: 
        \STATE $\boldsymbol{U}_0=\mathrm{orth}(\boldsymbol{D}(:,1:d))$
        \STATE $\boldsymbol{V}_0=\boldsymbol{D}^{*}\boldsymbol{U}_{0}$
        \STATE $\boldsymbol{B}_0=\mathrm{zeros}(N_s,N_t)+j\mathrm{zeros}(N_s,N_t)$
        \STATE $k=1$
        \WHILE{$k\leq $(maximum iteration) } 
            \STATE Update $\boldsymbol{W}_{b,k}$ using Eq.(\ref{w_b})
            \STATE Update $\boldsymbol{B}_{k}$ using Eq.(\ref{B_{k+1}})
            \STATE Update $\boldsymbol{V}_{k}$ using Eq.(\ref{V_{k+1}})
            \STATE Update $\boldsymbol{U}_{k}$ using Eq.(\ref{U_{k+1}})
            \STATE Update $\boldsymbol{W}_{c,k}$ using Eq.(\ref{w_c})
           \IF{$\Big(\dfrac{{\lVert\boldsymbol{B}_k+\boldsymbol{C}_k-\boldsymbol{B}_{k-1}-\boldsymbol{C}_{k-1}}\rVert_{F}^{2}}{ {\lVert \boldsymbol{B}_{k-1}+\boldsymbol{C}_{k-1} \rVert}_{F}^{2}}\Big) < \epsilon_0$} 
		      \STATE \textbf{break}
		      \ENDIF
        \ENDWHILE
        
        \RETURN $\boldsymbol{B}\leftarrow \boldsymbol{B}_{k}$ 
    \end{algorithmic}
\end{algorithm}

To properly unfold the rPCA clutter filter into an unsupervised network, new relaxations to the nuclear norm and the $l_1$-norm are needed. Fortunately, Daubechies \textit{et al}.~\cite{daubechies2010iteratively} proposed and proved that a solution $\boldsymbol{x}^*$ of $l_1$-norm coincides with a unique sparse-recovery solution $\boldsymbol{x}$ of an iteratively reweighted least squares (IRLS) sparse minimization ${\lVert\boldsymbol{x} \odot \boldsymbol{w}^{\frac{1}{2}}\rVert}_{2}^{2}$, where $\boldsymbol{w}_j = {\lvert \boldsymbol{x}^*_j \rvert}^{-1}$ and $\odot$ denotes Hadamard product. Motivated by~\cite{daubechies2010iteratively}, Giampouras \textit{et al}.~\cite{giampouras2018alternating} extended IRLS to low-rank minimization. They proposed and proved that the nuclear norm ${\lVert \boldsymbol{X} \rVert}_{*}$ has a tight upper bound $\Big({\lVert\boldsymbol{U}\boldsymbol{W}^{\frac{1}{2}}\rVert}_{F}^{2}+{\lVert\boldsymbol{V}\boldsymbol{W}^{\frac{1}{2}}\rVert}_{F}^{2}\Big)$, where $\boldsymbol{X} = \boldsymbol{U} \boldsymbol{V}^*$ and $\boldsymbol{W}$ is the corresponding weights defined as $\boldsymbol{W} =\operatorname{diag}\bigg( \Big({\lVert \boldsymbol{U}(:,1) \rVert}_{2}^{2}+{\lVert \boldsymbol{V}(:,1) \rVert}_{2}^{2}+\epsilon \Big)^{\frac{\rho}{2}-1},\dots,\Big({\lVert \boldsymbol{U}(:,d) \rVert}_{2}^{2}+{\lVert \boldsymbol{V}(:,d) \rVert}_{2}^{2}+\epsilon \Big)^{\frac{\rho}{2}-1}\bigg)$, where $0<\rho \leq 1$. ${\lVert \cdot \rVert}_F$ denotes the matrix Frobenius norm. Therefore, the low-rank $\boldsymbol{C}$ can be equivalently formulated as $\boldsymbol{C}=\boldsymbol{U}\boldsymbol{V}^{\ast}$, where $\boldsymbol{U}\in \mathbb{C}^{N_s, d}$ is the basis that spans the subspace of $\boldsymbol{C}$. $\boldsymbol{V}\in \mathbb{C}^{N_t, d}$ acts as the coordinate coefficients of $\boldsymbol{C}$ under $\boldsymbol{U}$. As a result, Eq.~(\ref{rpca}) is rewritten as 
\begin{equation}
    \boldsymbol{D}=\boldsymbol{U}\boldsymbol{V}^{\ast}+\boldsymbol{B}+\boldsymbol{N}
\end{equation}

\noindent where the superscript $\ast$ means conjugate transpose. $d$ is the inner dimension of $\boldsymbol{C}$ with $r < d\ll \min{(N_s,N_t)}$, where $r$ is the true rank of $\boldsymbol{C}$. The inner dimension $d$ is larger than $r$ because the true rank of tissue signals is not exactly known. The tissue basis should be designed with redundancy to deal with the complex imaging conditions. With the IRLS approach, the baseline of U\textsuperscript{2}-rPCA is formulated as~\cite{rontogiannis2020online,OnlineChen} 
\begin{equation}
\begin{aligned}
    \min_{\substack{\boldsymbol{U},\boldsymbol{V},\boldsymbol{B},\\ \boldsymbol{W_c},\boldsymbol{W_b}}}\, & \lambda_c \Big({\lVert\boldsymbol{U}\boldsymbol{W}_{c}^{\frac{1}{2}}\rVert}_{F}^{2}+{\lVert\boldsymbol{V}\boldsymbol{W}_{c}^{\frac{1}{2}}\rVert}_{F}^{2}\Big)+
    \lambda_b {\lVert\boldsymbol{B}\odot \boldsymbol{W}_{b}^{\frac{1}{2}}\rVert}_{F}^{2}\\
    \text{s.t.}\, &\boldsymbol{U}\boldsymbol{V}^{\ast} + \boldsymbol{B} = \boldsymbol{D}
    \label{IRLS_LOSS}
\end{aligned}
\end{equation}
\noindent where $\lambda_c$ and $\lambda_b$ are the penalty coefficients for the low-rank and the sparse regularization, respectively. $\boldsymbol{W}_c\in \mathbb{R}^{d, d}$ are the weights for the low-rank term:
\begin{equation}
\begin{aligned}
    \boldsymbol{W}_c &=\operatorname{diag}\bigg( \Big({\lVert \boldsymbol{U}(:,1) \rVert}_{2}^{2}+{\lVert \boldsymbol{V}(:,1) \rVert}_{2}^{2}+\epsilon \Big)^{-\frac{1}{2}},\dots,\\
    &\Big({\lVert \boldsymbol{U}(:,d) \rVert}_{2}^{2}+{\lVert \boldsymbol{V}(:,d) \rVert}_{2}^{2}+\epsilon \Big)^{-\frac{1}{2}}
    \bigg),\label{w_c}
\end{aligned}
\end{equation}

\noindent where $\operatorname{diag}(\cdot)$ transfers a given vector into a diagonal matrix. $\epsilon$ is a small positive constant to avoid singularity. According to~\cite{daubechies2010iteratively} and~\cite{rontogiannis2020online}, each element of the weight matrix $\boldsymbol{W}_b \in \mathbb{R}^{N_s, N_t}$ for the sparse term is defined as
\begin{equation}
    \boldsymbol{W}_b(i,j)=(\boldsymbol{B}(i,j)^{2}+\epsilon)^{-\frac{1}{2}}\label{w_b}
\end{equation}

The constraint of Eq.~(\ref{IRLS_LOSS}) can be substituted with a data-consistency term, resulting in the following quadratic minimization problem:
\begin{equation}
\begin{aligned}
    \min_{\boldsymbol{U},\boldsymbol{V},\boldsymbol{B},\boldsymbol{W_c},\boldsymbol{W_b}}\frac{1}{2}{\lVert \boldsymbol{D}-\boldsymbol{U}\boldsymbol{V}^{\ast}-\boldsymbol{B}\rVert }_{F}^{2}\\
    +\lambda_c \Big({\lVert\boldsymbol{U}\boldsymbol{W}_{c}^{\frac{1}{2}}\rVert}_{F}^{2}+{\lVert\boldsymbol{V}\boldsymbol{W}_{c}^{\frac{1}{2}}\rVert}_{F}^{2}\Big)+
    \lambda_b {\lVert\boldsymbol{B}\odot \boldsymbol{W}_{b}^{\frac{1}{2}}\rVert}_{F}^{2}\
    \label{IRLS_LOSS2}
\end{aligned}
\end{equation}

Eq.~(\ref{IRLS_LOSS2}) is solved by alternately updating $\boldsymbol{B}_{k}$, $\boldsymbol{V}_{k}$, $\boldsymbol{U}_{k}$, and the weight matrices until convergence. Here $k$ denotes the index of iterations. The specific updating formulations are clarified as follows:


\begin{small}
\begin{align}
\boldsymbol{B}_{k} &= (\boldsymbol{D}-\boldsymbol{U}_{k-1}\boldsymbol{V}_{k-1}^{\ast})\oslash(1+2\lambda_{b}\boldsymbol{W}_{b,k}) \label{B_{k+1}} \\
\boldsymbol{V}_{k}& =(\boldsymbol{D}-\boldsymbol{B}_{k})^{\ast}\boldsymbol{U}_{k-1}(\boldsymbol{U}_{k-1}^{\ast}\boldsymbol{U}_{k-1}+2\lambda_{c}\boldsymbol{W}_{c,k-1})^{-1} \label{V_{k+1}}  \\
\boldsymbol{U}_{k} &=(\boldsymbol{D}-\boldsymbol{B}_{k})^{\ast}\boldsymbol{V}_{k}(\boldsymbol{V}_{k}^{\ast}\boldsymbol{V}_{k}+2\lambda_{c}\boldsymbol{W}_{c,k-1})^{-1} \label{U_{k+1}}
\end{align}
\end{small}

\noindent where $\oslash$ denotes element-wise division. In each iteration, $\boldsymbol{W}_{b,k}$ is updated according Eq.~(\ref{w_b}) before $\boldsymbol{B}_{k}$. $\boldsymbol{W}_{c,k}$ is updated using Eq.~(\ref{w_c}) following $\boldsymbol{U}_{k}$.
\textbf{Algorithm~\ref{alg:IRLS}} summarizes the realization of IRLS-rPCA clutter filtering. As illustrated in Eq. (\ref{IRLS_LOSS2}), all the terms are formulated with the Frobenius norm, which is mathematically equivalent to the $l_2$-norm, or more specifically, the MSE loss function for deep learning. The good properties of continuity and differentiability of the Frobenius norm can facilitate the unsupervised training of U\textsuperscript{2}-rPCA.

\subsection{U\textsuperscript{2}-rPCA Architecture}
Fig.~\ref{fig:model} shows the architecture of U\textsuperscript{2}-rPCA. The IRLS-rPCA baseline is unfolded into a succession of $K$ layers. The number of layers, $K$, is like the number of iterations for convergence in \textbf{Algorithm~\ref{alg:IRLS}}. Therefore, this hyperparameter is task-driven and should be predefined according to different imaging scenarios. The calculations in each layer are similar to a loop in \textbf{Algorithm ~\ref{alg:IRLS}}. Specifically, in the $k$-th layer, the blood flow $\boldsymbol{B}$, the tissue basis $\boldsymbol{U}$, and the tissue coordinate coefficients $\boldsymbol{V}$ are updated based on the outputs of the $(k-1)$-th layer and put forward to the $(k+1)$-th layer. $\boldsymbol{W}_{b}$ is deterministic, whereas $\boldsymbol{W}_{c}$ is set as learnable. Furthermore, a sparse-enhancement unit (SEU) right after the update of $\boldsymbol{B}$ is plugged into each layer. 

\subsubsection{$\boldsymbol{B}$ Updating Unit}
As illustrated in Fig.~\ref{fig:model}(a), this unit first takes $\boldsymbol{B}_{k-1}$ to update $\boldsymbol{W}_{b,k}$ according to Eq.~(\ref{w_b}). Then, $\boldsymbol{U}_{k-1}$ and $\boldsymbol{V}_{k-1}$ of the $(k-1)$-th layer, $\boldsymbol{W}_{b,k}$ of this layer, and the raw sequence $\boldsymbol{D}$ are put together to update $\boldsymbol{B}_{k}$. Similar to Eq.~(\ref{B_{k+1}}), the formulation of the $\boldsymbol{B}$ updating unit is 
\begin{equation}
    \boldsymbol{B}_{k}=(\boldsymbol{D}-\boldsymbol{U}_{k-1}\boldsymbol{V}_{k-1}^{\ast})\oslash(1+2\lambda_{b,k}\boldsymbol{W}_{b,k})
    \label{B_{k+1}2}
\end{equation}

The difference between Eq.~(\ref{B_{k+1}2}) and Eq.~(\ref{B_{k+1}}) is that the penalty coefficient $\lambda_{b,k}$ becomes a learnable parameter. It should be noted that $\boldsymbol{W}_{b,k}$ is deterministic according to Eq.~(\ref{w_b}) rather than being set as learnable parameters. This is because Eq.~(\ref{w_b}) is a sparsity constraint with rigorous mathematical proofs. The deterministic $\boldsymbol{W}_{b,k}$ helps the U\textsuperscript{2}-rPCA network maintain its intrinsic sparse regularization.  

%
%
%
%
%
\subsubsection{Sparse Enhancement Unit (SEU)} 
The unsupervised manner of U\textsuperscript{2}-rPCA means that the network cannot learn the feature differences between tissue and blood flow from referenced or ground-truth datasets. The consequence is that the performance of the unfolded version may be very close to that of the IRLS-rPCA baseline. To strengthen the capability of U\textsuperscript{2}-rPCA in extracting the features of sparse blood flow signals in the spatial domain, we added a plug-in SEU module after the $\boldsymbol{B}$ updating unit. SEU is composed of a dual-frame U-Net, which has been applied to recover high-frequency edges in sparse-view CT~\cite{han2018framing} and unsupervised video foreground separation~\cite{takeda2024unsupervised}. Fig.~\ref{fig:model}(b) shows that the dual-frame U-Net is a symmetrical U-shaped network. In this module, the features obtained by upsampling in the decoder will be subtracted from the features downsampled in the encoder at the same scales. As a result, more refined sparse features of the blood flow can be extracted and fed into the following units. The output of SEU is formulated as 
\begin{equation}
    \boldsymbol{Z}_{k}=\mathrm{SEU}_{\theta,k}(\boldsymbol{B}_{k})
    \label{Z_{k+1}2}
\end{equation}

\noindent where $\operatorname{SEU}(\cdot)$ denotes the operation of the SEU module. ${\theta,k}$ are the learnable parameters of the dual-frame U-Net in the $k$-th layer. As a plug-in module, the inputs and outputs of this module are the features representing the sparse blood flow signals with identical dimensions. Either the inputs or the outputs of this module can be passed to the following unit. It should be noted that the dual-frame U-Net is realized in a complex network because the entire workflow of U\textsuperscript{2}-rPCA is implemented on the complex data.
\subsubsection{$\boldsymbol{V}$ and $\boldsymbol{U}$ Updating Units}
The unfolding operation turns $\lambda_c$ and $\boldsymbol{W}_c$ in Eq.~(\ref{IRLS_LOSS2}) into learnable parameters. Notably, $\lambda_c$ can be integrated into the learning of $\boldsymbol{W}_c$. The advantage is that U\textsuperscript{2}-rPCA becomes free of $\lambda_c$, which originally requires extensive fine-tuning for the success of the IRLS-rPCA baseline.
With this improvement, the $\boldsymbol{V}$ updating unit takes the output of the SEU $\boldsymbol{Z}_k$, the previous tissue basis $\boldsymbol{U}_{k-1}$, and the raw sequence $\boldsymbol{D}$ as input. The updated $\boldsymbol{V}_{k}$ is defined as 
\begin{equation}
    \boldsymbol{V}_{k}=(\boldsymbol{D}-\boldsymbol{Z}_{k})^{\ast}\boldsymbol{U}_{k-1}
    (\boldsymbol{U}_{k-1}^{\ast}\boldsymbol{U}_{k-1}+\boldsymbol{W}_{c,k})^{-1}\\
    \label{V_{k+1}2}
\end{equation}

Similarly, $\boldsymbol{V}_{k}$ and other necessary components are fed into the following $\boldsymbol{U}$ updating unit for the update of $\boldsymbol{U}_k$:
\begin{equation}
    \boldsymbol{U}_{k}=(\boldsymbol{D}-\boldsymbol{Z}_{k})^{\ast}\boldsymbol{V}_{k}
    (\boldsymbol{V}_{k}^{\ast}\boldsymbol{V}_{k}+\boldsymbol{W}_{c,k})^{-1}
    \label{U_{k+1}2}
\end{equation}
Compared to the IRLS-rPCA baseline, the only hyperparameter in this unit is the inner dimension of the tissue basis, $d$. This hyperparameter limits the dimensions of $\boldsymbol{U}_k$ and $\boldsymbol{V}_k$ so that U\textsuperscript{2}-rPCA can maintain its intrinsic low-rank regularization to the tissue signals.

\subsection{Loss Function}
Eq.~(\ref{IRLS_LOSS2}) shows that IRLS-rPCA uses the Frobenius norm to model all the optimized terms. The definition of the matrix Frobenius norm is mathematically equivalent to the vector $l_2$-norm. Therefore, it is straightforward and rational to employ MSE as the loss function to train the network. Furthermore, we refer to the deep supervised scheme and apply unsupervised MSE loss functions to the outputs of all the layers. The final loss function of U\textsuperscript{2}-rPCA is defined as 
\begin{equation}
    Loss=\frac{1}{K}\sum_{k=1}^{K}{\lVert \boldsymbol{D}-\boldsymbol{B}_k-\boldsymbol{U}_{k}\boldsymbol{V}_{k}^{\ast}\rVert}_{F}^{2}
    \label{loss}
\end{equation}

\noindent where $K$ is the number of layers. It is worth mentioning that the low-rankness of $\boldsymbol{U}_k \boldsymbol{V}_k^{\ast}$ and the sparsity of $\boldsymbol{B}_k$ can be promoted during the layer-wise alternate updates. Therefore, there is no need to add extra low-rank or sparse regularization to the loss function.

\section{EXPERIMENTAL SETTINGS}
\subsection{In-Silico Experiments}

\begin{figure}[t]
\centering
    \includegraphics[width=0.5\textwidth]{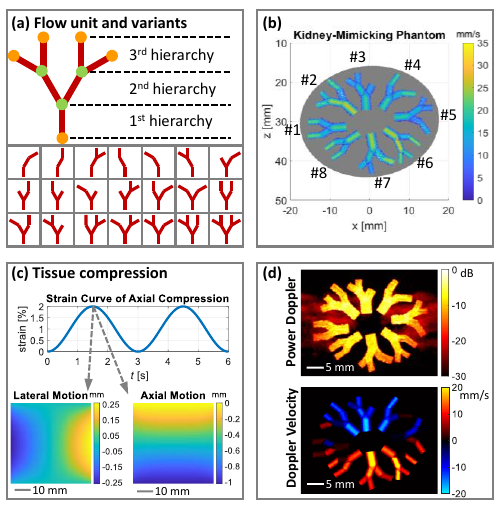}
    \caption{Configurations of the \textit{in-silico} kidney-mimicking phantom. (a) illustrates the basic flow unit and its variants. (b) is the geometry of this phantom and the velocities in the flow units. (c) shows the finite-element-analysis (FEA)-based axial-compression strain curve and the lateral and axial displacements applied to the phantom. (d) presents the ground-truth power Doppler image and Doppler velocity image.}
    \label{fig:simulation_config}
\end{figure}

Fig.~\ref{fig:simulation_config} illustrates the simulation configurations. Inspired by the kidney microvasculature, we constructed a kidney-mimicking phantom that comprised eight multi-hierarchy flow units distributing radially in an oval-shaped region (see Fig.~\ref{fig:simulation_config}(b)). The complete flow unit was composed of three hierarchies, each of which included 1, 2, and 4 sub-branches, respectively (see Fig.~\ref{fig:simulation_config}(a)). Fig.~\ref{fig:simulation_config}(a) also lists all the sub-structures of the unit that contain three hierarchies. Eight sub-structures were randomly selected with repeatability to form the numerical kidney phantom in Fig.~\ref{fig:simulation_config}(b).

Randomness was also applied to the parameters of the flow units. Table~\ref{tab:parameters} shows that the lengths ($L_1$, $L_2$, and $L_3$) and steering angles ($S_1$, $S_2$, and $S_3$) of all the sub-branches, and the radius of the first sub-branch ($R_1$) were randomly specified in the given ranges. The radii of the following sub-branches ($R_2$ and $R_3$) were determined accordingly to maintain the smoothness of the unit geometry. The maximal velocity in the first sub-branch of each unit was randomly set between 20 mm/s and 30 mm/s. The flow dynamics of the other sub-branches of each flow unit were calculated according to the assumptions of parabolic laminar flow and fluid incompressibility. 

The tissue region, i.e., the gray area in Fig.~\ref{fig:simulation_config}(b), was generated by rotating a thin oval-shaped disk with a long axis of 35 mm and a short axis of 28 mm. The rotating angle was set as a random value in the range of [$-10^{\circ}$, $10^{\circ}$].
Respiration-mimicking tissue motions were applied to the numerical phantom (Fig.~\ref{fig:simulation_config}(c)). The applied axial and lateral displacements that mimic the compression-recovery tissue motion cycles were simulated via finite element analysis (FEA) in FEBio~\cite{FEBio}. The maximal axial strain was $2\%$. The motion cycle rate was 1/3 Hz, mimicking the normal human respiration rate of 20 beats per minute.


A plane wave compounding sequence was simulated in Field II~\cite{FieldII} to image the numerical phantom. Specifically, a L10-5 linear array with a center frequency of 7.5 MHz was used to transmit 5 plane waves with steering angles of [$-10^{\circ}$, $-5^{\circ}$, $0^{\circ}$, $5^{\circ}$, $10^{\circ}$]. The pulse repetition frequency (PRF) was 5 kHz, resulting in a frame rate of 1 kHz. The acquisition time was 6 seconds, containing a total number of 6,000 frames. Gaussian white noise with an SNR of 25 dB was added to the beamformed data. The first 5,400 frames were used to train U\textsuperscript{2}-rPCA and 3D-Res-UNet, whereas the remaining 600 frames were used for testing.

\begin{figure}[t]
\centering
    \includegraphics[width=0.5\textwidth]{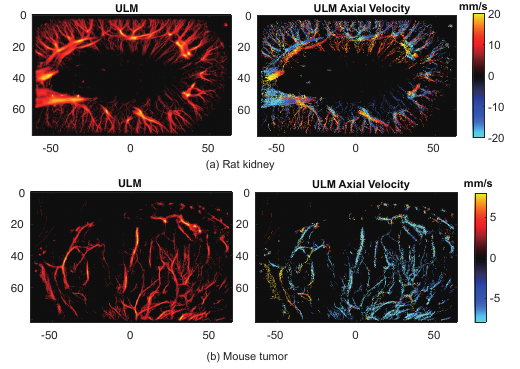}
    \caption{ULM density maps and axial velocity components (references of the Doppler velocity estimations) of (a) the kidney dataset and (b) the tumor dataset obtained from the PALA scripts and data.}
    \label{fig:ULM_pd_dv}
\end{figure}

\begin{table}[t]
\centering
\caption{Simulation Parameters of the Flow Units}
\begin{tabular}{>{\centering\arraybackslash}m{1.5cm} >{\centering\arraybackslash}m{1cm} >{\centering\arraybackslash}m{5cm}}
\toprule
\textbf{Parameter} & \textbf{Hierarchy} & \textbf{Range of value} \\
\midrule
$\textit{L}$ & $1$ & $3.5 + [-0.35, 0.35] ~\text{mm}$ \\
& $2$ & $3.5 + [-0.35, 0.35] ~\text{mm}$ \\
& $3$ & $3.5 + [-0.35, 0.35] ~\text{mm}$ \\
\midrule
$\textit{S}$ & $1$ & $[-10^\circ, 10^\circ] $ \\
& $2$ & $\pm 30^\circ \pm [-2^\circ, 2^\circ] $ \\
& $3$ & $60^\circ \pm [-3^\circ, 3^\circ] \text{and} [-3^\circ, 3^\circ] $ \\
\midrule
$\textit{R}$ & $1$ & $1.25 \pm [-0.125,0.125] ~\text{mm} $ \\
 & $2$ & Depends on $R_1$  \\
  & $3$ & Depends on $R_2$  \\
  \midrule
  $_{\mathnormal{V_{max}}}$ & $1$ & $ [20,30] ~\text{mm/s} $ \\
 & $2$ & Depends on $V_{max1}$ \\
  & $3$ & Depends on $V_{max1}$ and $V_{max2}$ \\
\bottomrule
\end{tabular}
\label{tab:parameters}
\end{table}

\subsection{In-Vivo Experiments}
%

We used the public PALA datasets provided in~\cite{heiles2022performance} to validate the effectiveness of  U\textsuperscript{2}-rPCA. PALA provides four \textit{in-vivo} contrast-enhanced datasets including rat brain, rat brain bolus, rat kidney, and mouse tumor. We chose the kidney and the tumor datasets because the selected objects are more likely to suffer from complex non-rigid motions caused by respiration and heartbeat. Both datasets were acquired with a five-angle plane wave imaging sequence running on a 15-MHz high-frequency probe. Contrast agents were used to enhance the visibility of blood flow. The image size was 78 $\times$ 128 for the kidney and 83 $\times$ 128 for the tumor. 

Fig.~\ref{fig:ULM_pd_dv} presents the ULM density maps and the axial-velocity maps of the kidney and tumor datasets, which provide references to power Doppler imaging and Doppler velocity estimation, respectively. The total number of frames is 171,360 at a frame rate of 1 kHz for the kidney and 30,000 at a frame rate of 500 Hz for the tumor. 
In both datasets, the first 14,400 frames were used to train the corresponding U\textsuperscript{2}-rPCA networks, the next 5,000 frames were used for validation, and the remaining frames were all used for testing.

\subsection{Comparison and Evaluation Metrics}
We adopted traditional SVD and rPCA methods, and one deep learning-based method (3D-Res-UNet) for comparison. The SVD-based method was realized by using the spatial similarity matrix for adaptive cutoff estimation~\cite{baranger2018adaptive}. 
The rPCA-based method was illustrated in \textbf{Algorithm~\ref{alg:IRLS}}. Choosing IRLS-rPCA for comparison also helps to validate the performance of U\textsuperscript{2}-rPCA relative to its baseline. For the IRLS-rPCA baseline, the inner dimension $d$ was fine-tuned from the singular rank when the singular value curve decreased to a threshold of 1\% of the maximal value. The final choices were 25 and 15 for the kidney and tumor datasets, respectively. We fine-tuned $\lambda_c$ as 0.01 and $\lambda_b$ as 0.0004 for the kidney dataset, and 0.001 and 0.0002 for the tumor dataset.
%

The 3D-Res-UNet method was proposed in~\cite{di2022deep}.
This network learns the mapping from input compounded frames to power Doppler and outputs power Doppler images directly. According to \cite{di2022deep}, 3D-Res-UNet was trained with power Doppler images obtained from the SVD-based filtering. Therefore, we strictly followed the structure of 3D-Res-UNet and trained it with the results obtained from the abovementioned SVD-based method. The training, validation, and testing datasets are the same as those applied to U\textsuperscript{2}-rPCA.



We used contrast-to-noise ratio~(CNR), signal-to-noise ratio~(SNR), and peak-to-side-level~(PSL) to quantitatively assess the power Doppler images:
\begin{equation}
    \mathrm{CNR} =10\mathrm{log}_{10}\Big(\frac{\mathrm{mean}(\mathbf{PW}_{b})-\mathrm{mean}(\mathbf{PW}_{c})}{\mathrm{std}(\mathbf{PW}_{c})}\Big)
\end{equation}
\begin{equation}
    \mathrm{SNR}=10\mathrm{log}_{10}\Big(\frac{\mathrm{mean}({\mathbf{PW}_{b}})}{\mathrm{std}(\mathbf{PW}_{c})}\Big) 
\end{equation}
\begin{equation}
    \mathrm{PSL}=10\mathrm{log}_{10}\Big(\frac{\mathrm{max}({\mathbf{PW}_{b}})}{\mathrm{mean}(\mathbf{PW}_{c})}\Big)
\end{equation}
where $\mathrm{mean}(\cdot)$, $\mathrm{std}(\cdot)$, and $\mathrm{max}(\cdot)$ calculate the average, the standard deviation, and the maximum of the power Doppler signals, respectively. The PSL is only used to validate the effectiveness of the plug-in SEU module. In addition, we used the correlation coefficient ($R^2$) between the Doppler velocities and the axial velocities of ULM to evaluate the quality of Doppler estimates.

\subsection{Implementation Details}
Two hyperparameters, $K$ and $d$ were predefined for U\textsuperscript{2}-rPCA. According to the parameter study presented in Sec. V-C, $K$ was set as 5 for the kidney dataset and 4 for the tumor dataset. $d$ was fixed to 40 and 38 for the kidney and tumor datasets, respectively. For the kidney dataset, $\lambda_b$ in each layer was initialized to 20. For the tumor dataset, this penalty coefficient was initialized to 10. 
In the simulation data, $K$, $d$, and $\lambda_b$ was set as 10, 10, and 6, respectively. 
The size of each training batch was $N_s \times N_{batch}$. $N_{batch}$ was set to 800 frames in the \textit{in-vivo} datasets and 200 frames in the \textit{in-silico} dataset.
For the realization of 3D-Res-UNet, all the parameters were set according to its default and recommended configurations.
Since the raw ultrasound sequences are in the in-phase and quadrature-phased (IQ) format, we expanded the traditional real-value network into a complex network for U\textsuperscript{2}-rPCA.
Experiments were conducted in a Linux environment using an Intel(R) Xeon(R) Silver 4210 CPU @ 2.20 GHz and an NVIDIA GeForce RTX 3090 GPU.
%
The network was trained using the Adam optimizer with an initial learning rate of 0.01 and 0.0001. An early stopping with a patience of 5 epochs was employed to avoid overfitting.

\begin{figure*}[t]
\centering
    \includegraphics[width=1\textwidth]{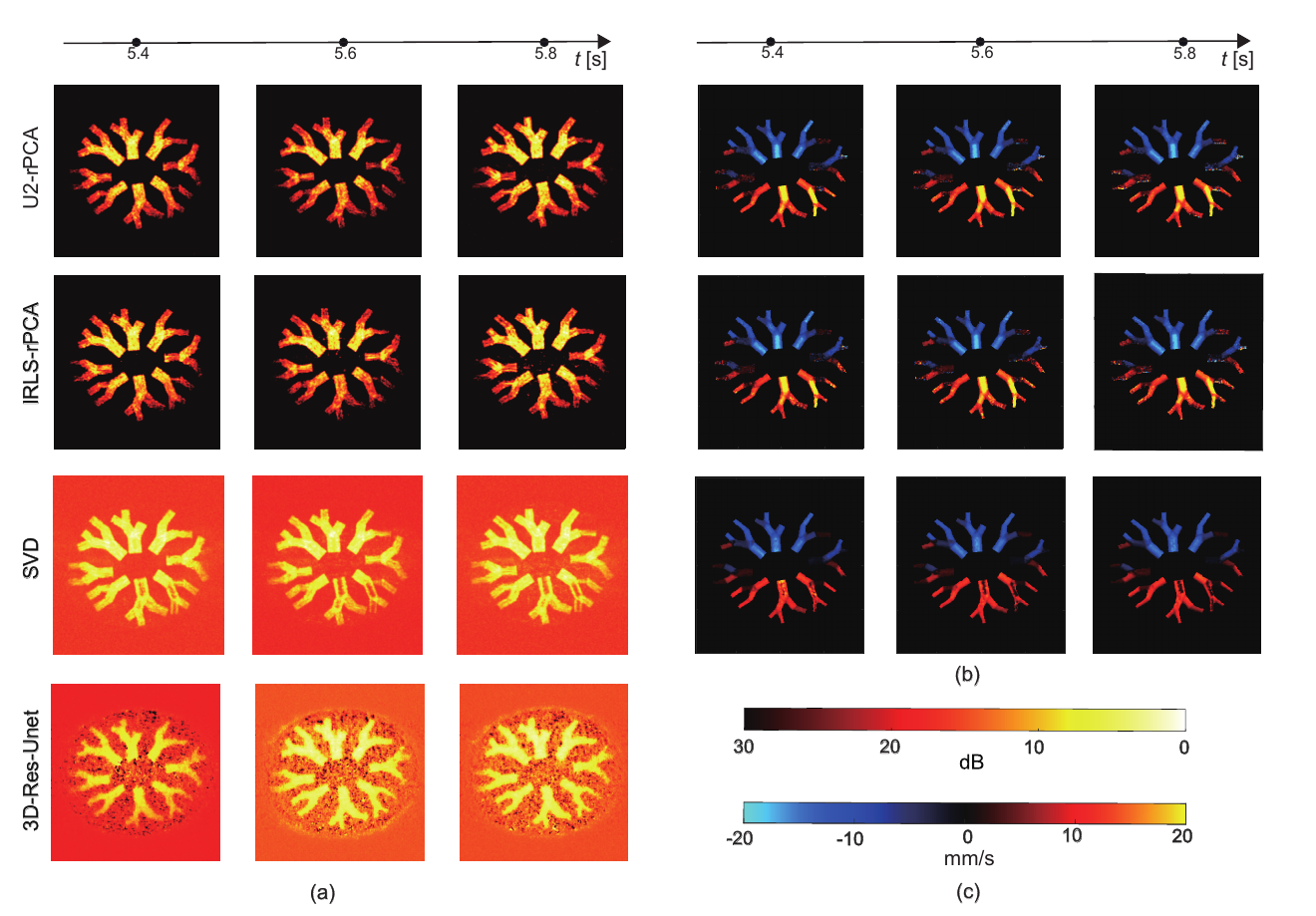}
    \caption{(a) Power Doppler images of the \textit{in-silico} kidney-mimicking phantom obtained by U\textsuperscript{2}-rPCA, IRLS-rPCA, SVD, and 3D-Res-UNet. (b) Doppler velocity images obtained by U\textsuperscript{2}-rPCA, IRLS-rPCA, and SVD. (c) The color bars illustrate the dynamic range and velocity range.}
    \label{fig:UPDI_simulations}
\end{figure*}

\begin{table}[t]
    \centering
     \caption{CNR of power Doppler and $R^2$ of Doppler velocity in the kidney-mimicking phantom}
     \resizebox{0.5\textwidth}{!}{ 
    \begin{tabular}{c|ccc|ccc}
    \toprule
    & \multicolumn{3}{c|}{CNR [dB]} &  \multicolumn{3}{c}{$R^2$} \\
    \midrule
     \diagbox{Method}{No. frame}& 1& 2& 3&  1& 2& 3\\
    \midrule
    U\textsuperscript{2}-rPCA  &  \textbf{27.95}& \textbf{27.79}& \textbf{26.68}& \textbf{0.9530}& \textbf{0.9002}& 0.8458\\
    IRLS-rPCA  &   16.07& 23.36& 19.48& 0.9476& 0.8975& \textbf{0.8538}\\
    SVD  &  18.78& 16.98& 16.45&  0.9471& 0.8722 & 0.8536\\ 
    3D-Res-UNet & 8.06& 10.56& 9.90& -& -& - \\
       \bottomrule
    \end{tabular}
    }
    \label{tab:simulation_metircs}
\end{table}

\section{RESULTS}

\subsection{Simulations}
Fig.~\ref{fig:UPDI_simulations} shows the power Doppler images and Doppler velocities of the \textit{in-silico} kidney-mimicking phantom obtained from the tested frames with an ensemble length of 200. Doppler velocity is unavailable for 3D-Res-UNet because this network only outputs power Doppler estimates. In a visual comparison, U\textsuperscript{2}-rPCA and IRLS-rPCA suppress tissue signals and noise to a lower level contrast to SVD and 3D-Res-UNet. U\textsuperscript{2}-rPCA and IRLS-rPCA achieve similar estimates of power Doppler and Doppler velocities. In addition, SVD loses some flow signals in the 6\textsuperscript{th} and 7\textsuperscript{th} flow units, leaving hollows in power Doppler images and the corresponding Doppler velocities. 

Table~\ref{tab:simulation_metircs} lists the CNRs of the power Doppler images and the correlations ($R^2$) between the Doppler velocities and the references. The tissue and flow region-of-interests (ROIs) are delineated from the references in Fig.~\ref{fig:simulation_config}. U\textsuperscript{2}-rPCA achieves the highest CNRs in all three power Doppler images, whereas IRLS-rPCA ranks the second best in the second and third frames. In terms of Doppler velocity, the tested methods achieve comparable correlations. Although SVD cannot correctly estimate Doppler velocities in the hollow regions in the 6\textsuperscript{th} and 7\textsuperscript{th} flow units, it provides smoother estimates in the 1\textsuperscript{st} and 5\textsuperscript{th} flow units than U\textsuperscript{2}-rPCA and IRLS-rPCA, where the flow directions are nearly parallel to the probe.

\begin{figure*}[!htb]
\centering
    \includegraphics[width=1\textwidth]{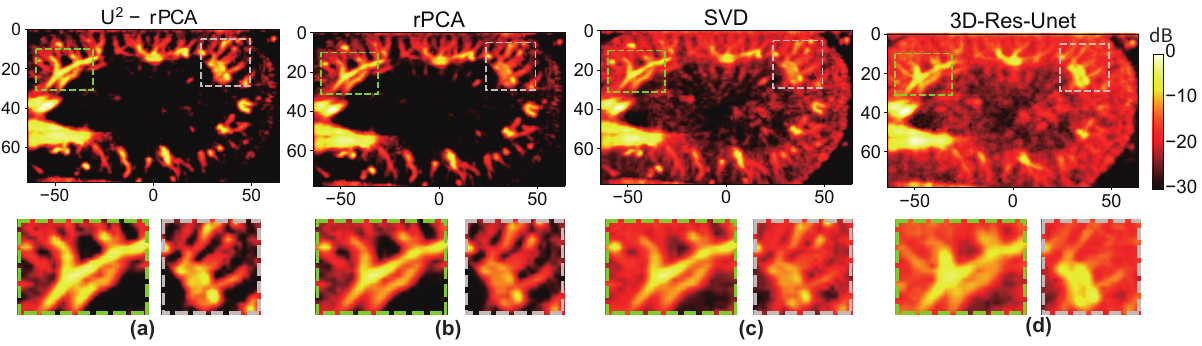}
    \caption{Power Doppler images of the kidney obtained by (a) U\textsuperscript{2}-rPCA, (b) IRLS-rPCA, (c) SVD, and (d) 3D-Res-UNet.}
    \label{fig:roi_enlarge_rat_kidney}
\end{figure*}

\begin{figure*}[!htb]
\centering
    \includegraphics[width=1\textwidth]{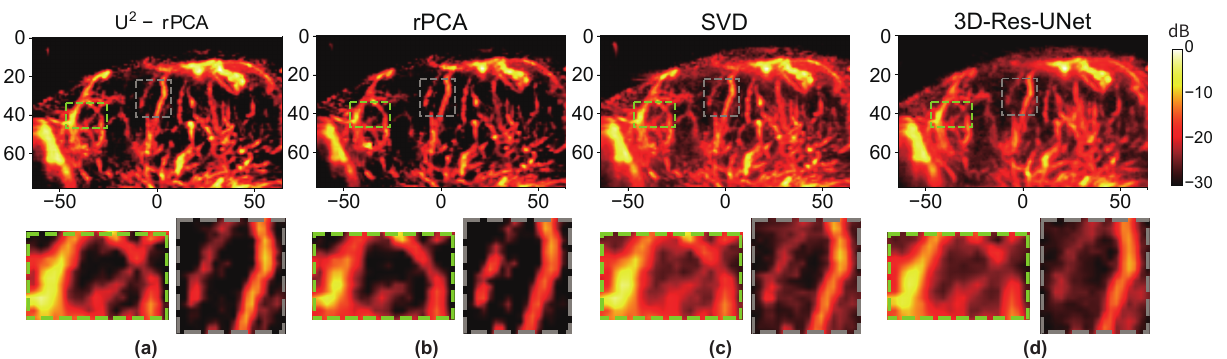}
    \caption{Power Doppler images of the tumor obtained by (a) U\textsuperscript{2}-rPCA, (b) IRLS-rPCA, (c) SVD, and (d) 3D-Res-UNet.}
    \label{fig:roi_enlarge_mouse_tumor}
\end{figure*}




\begin{table*}[!htb]
\centering
\caption{CNR and SNR on Two ROIs of the Two \textit{In-Vivo} Datasets (unit: dB).}
\label{table2}
\renewcommand{\arraystretch}{1.2} 
\setlength\tabcolsep{8pt} 
\begin{tabular}{c|cc|cc|cc|cc}
\toprule

ROI & \multicolumn{2}{c|}{ \includegraphics[width=0.1\linewidth]{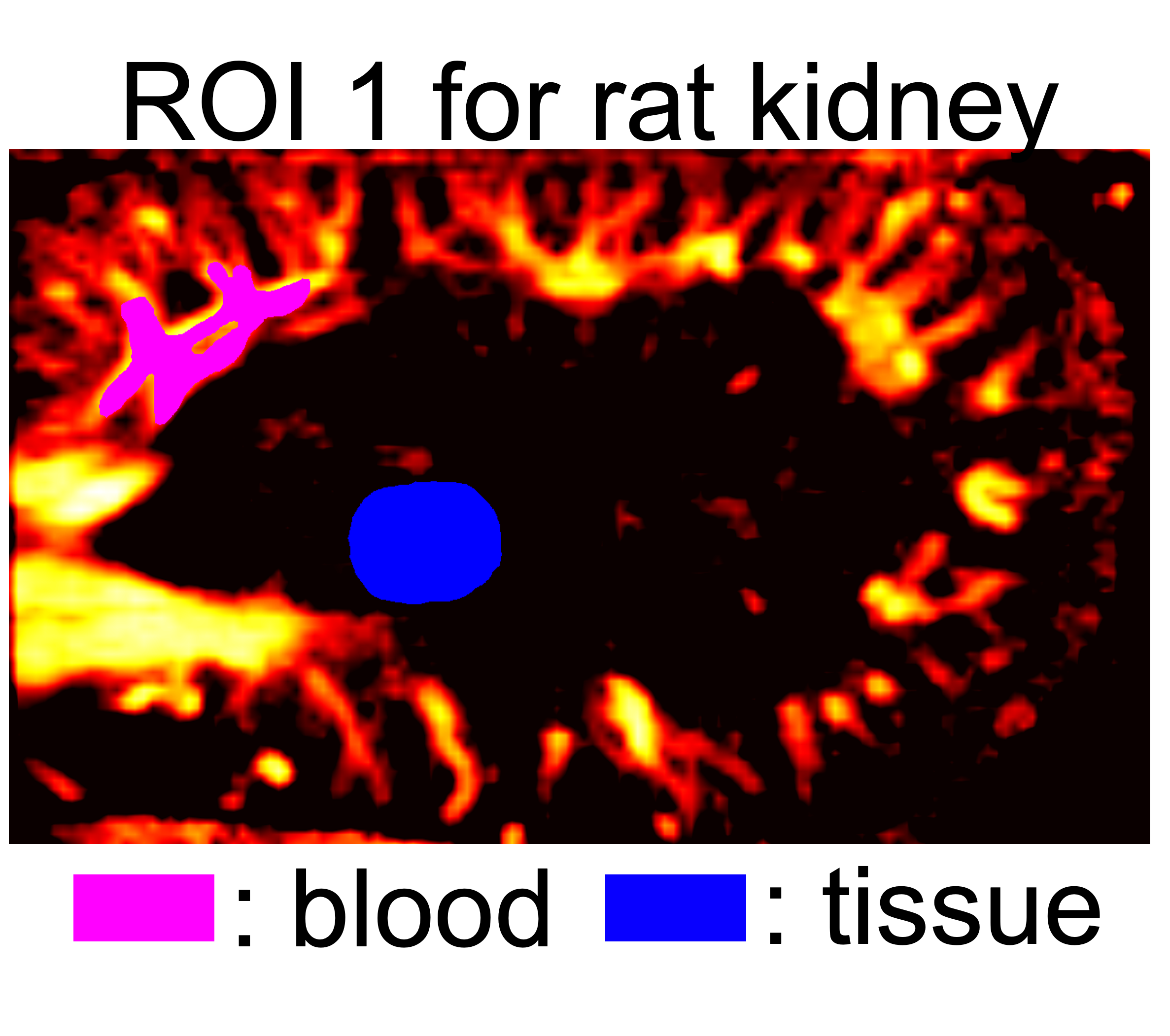}}& \multicolumn{2}{c|}{\includegraphics[width=0.1\linewidth]{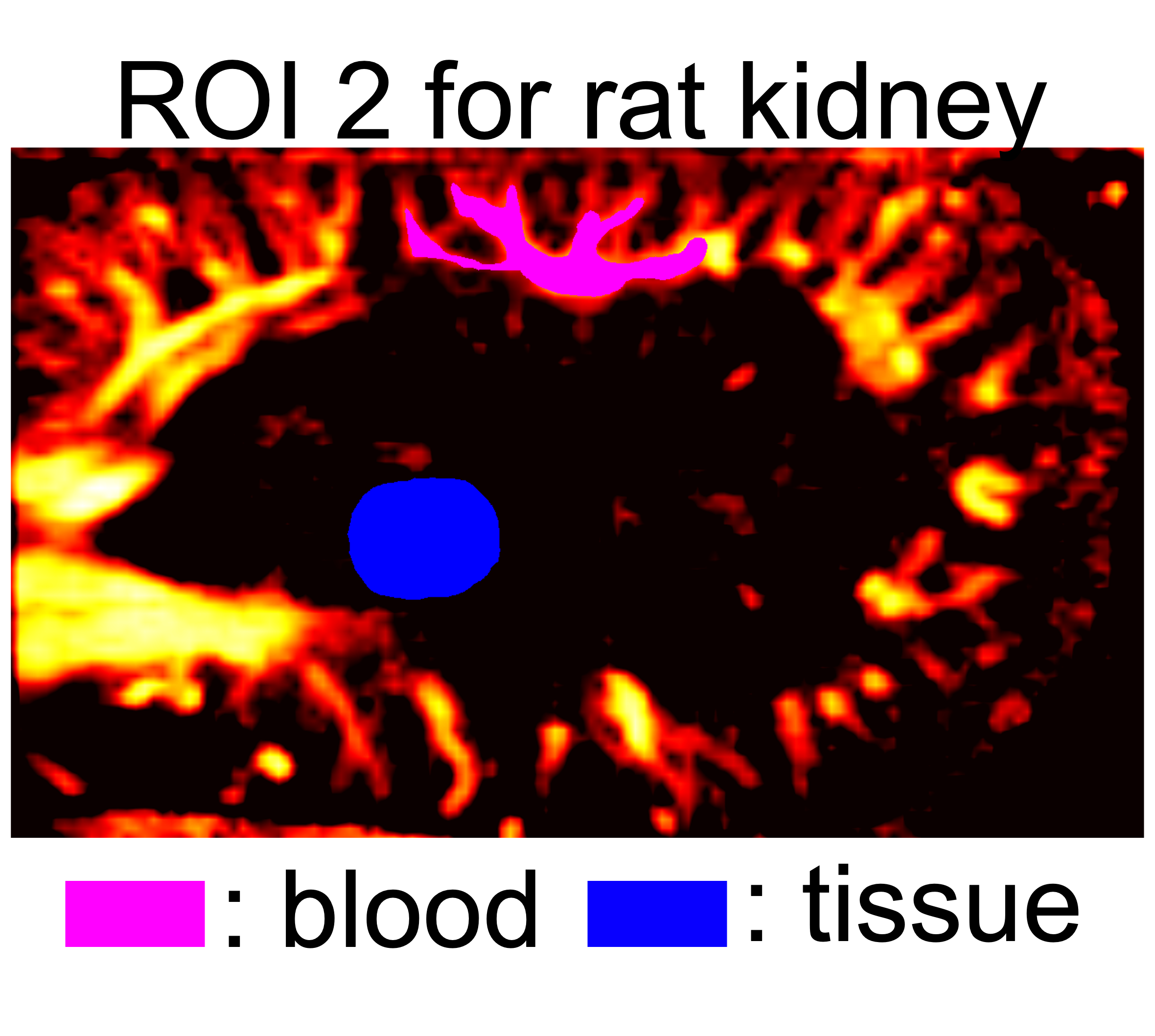}}&\multicolumn{2}{c|}{\includegraphics[width=0.1\linewidth]{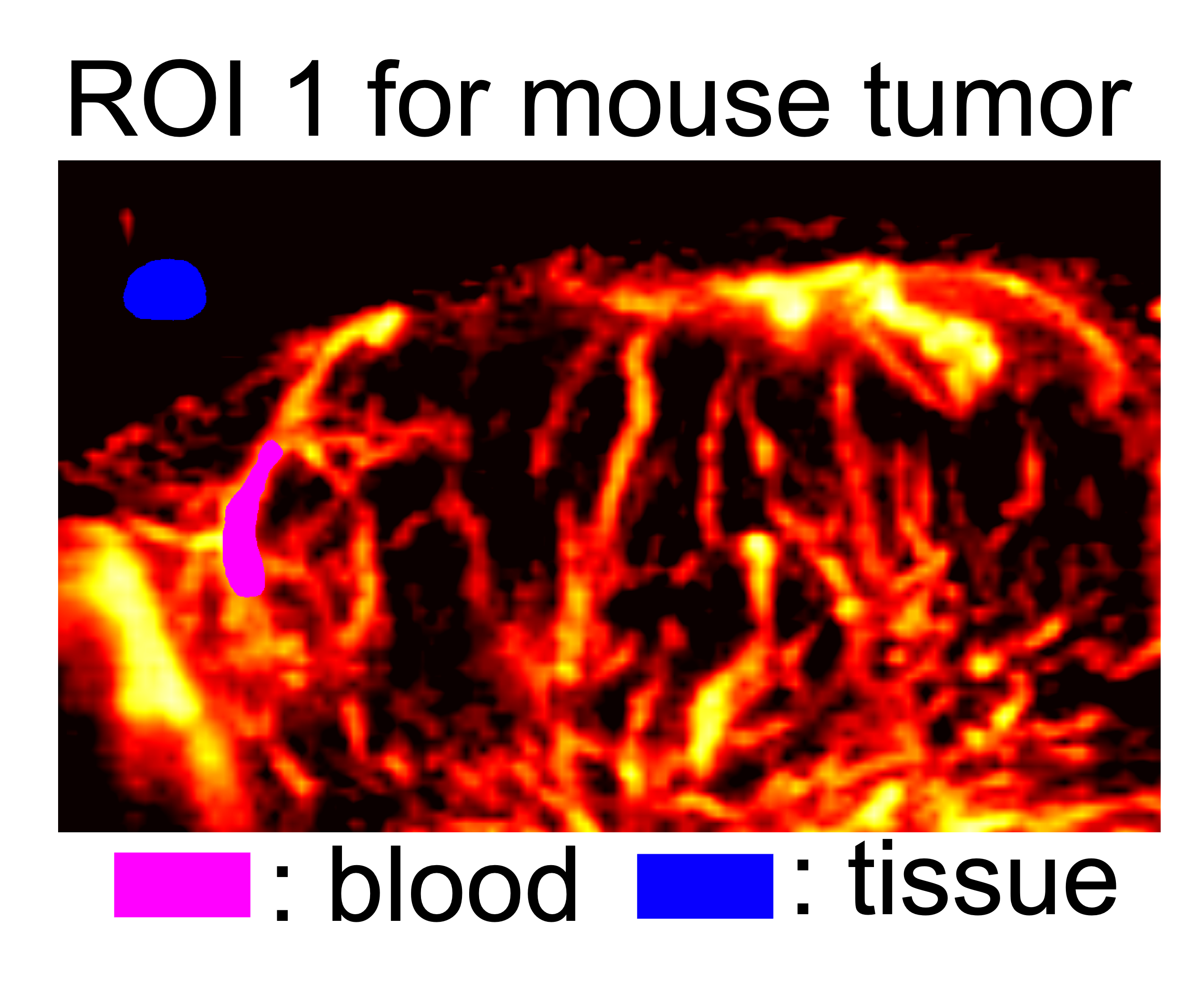}} 
& \multicolumn{2}{c}{\includegraphics[width=0.1\linewidth]{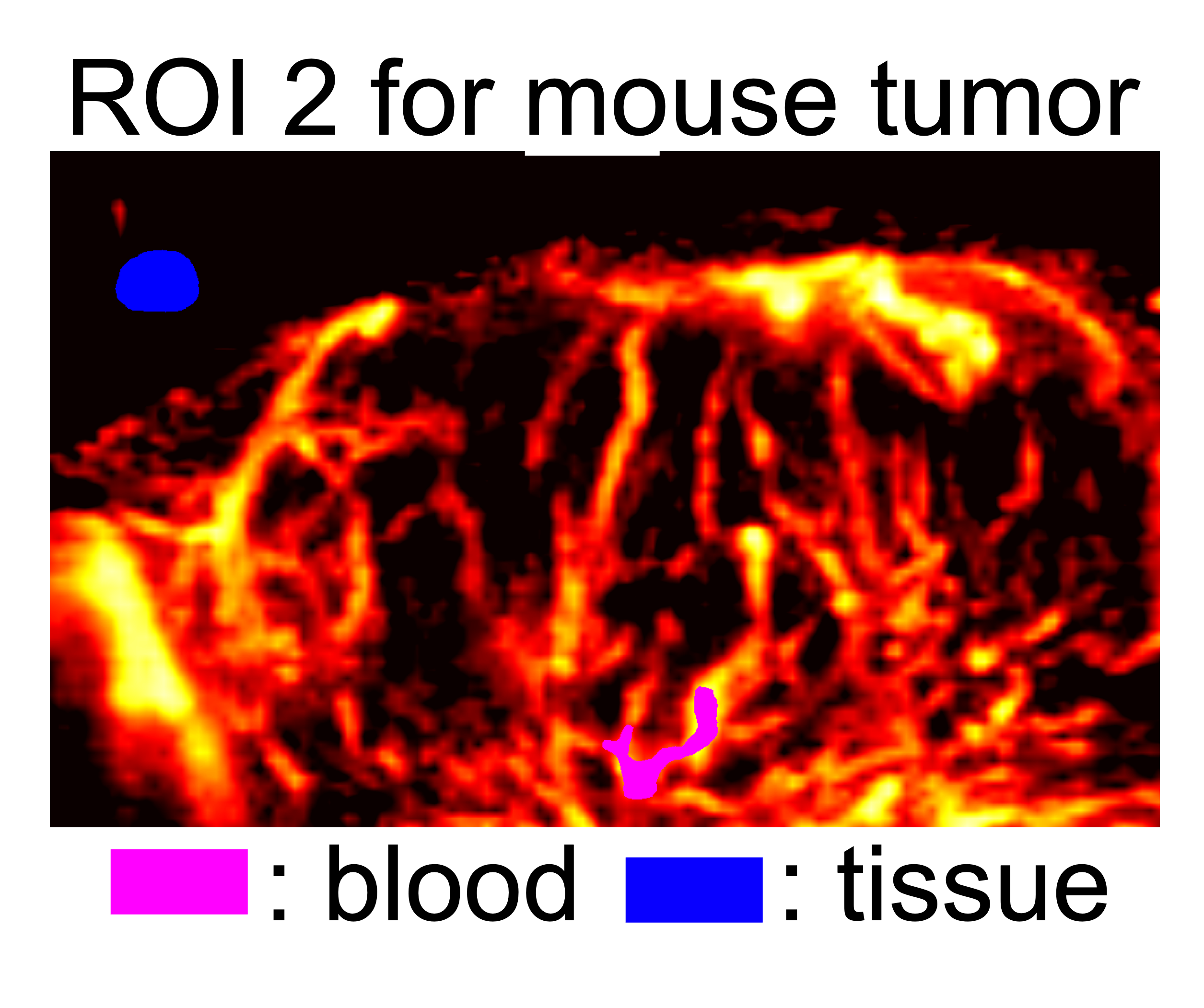}}\\

\midrule

Metric & CNR & SNR &CNR  &SNR  &CNR  &SNR &CNR & SNR\\
\midrule
U\textsuperscript{2}-rPCA& \textbf{27.91$\pm$2.19} & \textbf{27.91$\pm$2.18} & \textbf{26.42$\pm$1.99} & \textbf{26.43$\pm$1.99} & \textbf{34.25$\pm$7.26} & \textbf{34.27$\pm$7.23}  & \textbf{33.09$\pm$7.67} & \textbf{33.11$\pm$7.63} \\

IRLS-rPCA &  25.25$\pm$2.21 & 25.26$\pm$2.20 & 24.20$\pm$2.22 & 24.23$\pm$2.21 & 32.34$\pm$9.39 & 32.40$\pm$9.25 & 30.53$\pm$9.73 & 30.59$\pm$9.60 \\

SVD &  18.62$\pm$1.74 & 18.75$\pm$1.70 & 17.94$\pm$1.73 & 17.27$\pm$1.09 & 29.09$\pm$3.67 & 29.12$\pm$3.65 &  27.92$\pm$4.04 & 27.97$\pm$4.00\\

3D-Res-UNet &  18.57$\pm$1.36 & 18.71$\pm$1.33 & 17.21$\pm$1.61 & 17.42$\pm$1.57 & 30.90$\pm$1.26 & 30.94$\pm$1.25 & 30.22$\pm$2.12 & 30.27$\pm$2.10\\



\bottomrule
\end{tabular}
\label{cnr_snr}
\end{table*}

\begin{figure*}[!htb]
\centering
    \includegraphics[width=1\textwidth]{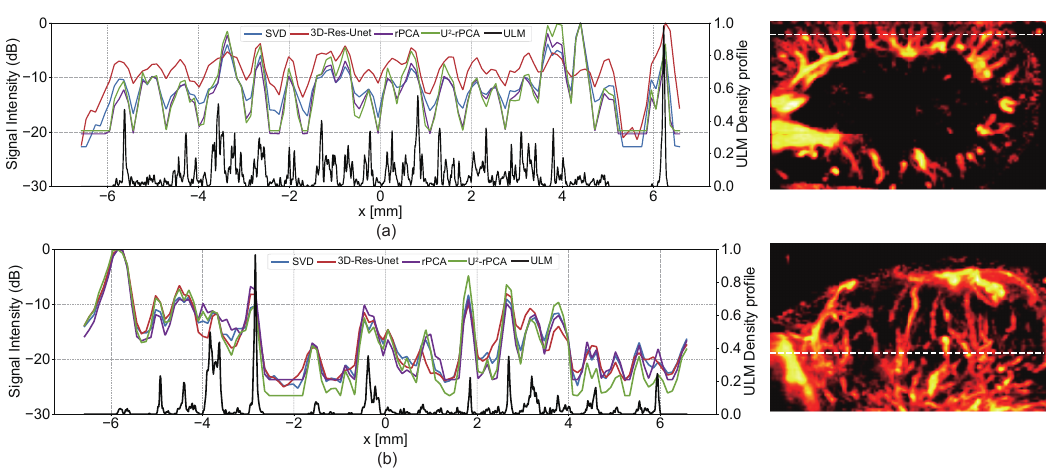}
    \caption{The normalized signal intensities (in dB) of the selected cross-section for (a) the kidney dataset and (b) the tumor dataset.}
    \label{fig:signal intensity}
\end{figure*}

\begin{figure*}[!htb]
\centering
    \includegraphics[width=1\textwidth]{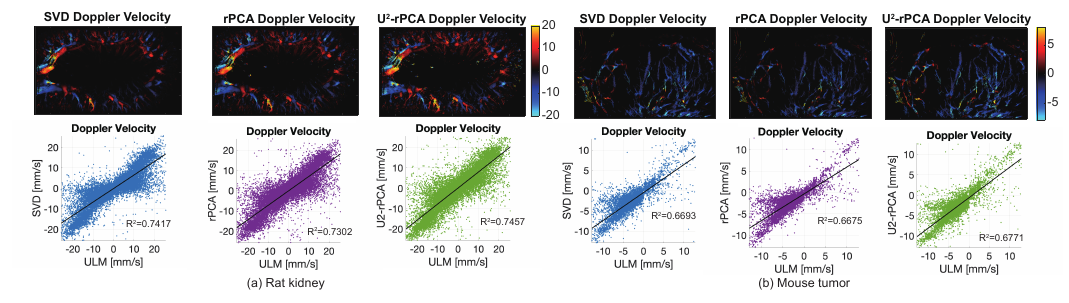}
    \caption{Doppler velocity images and the corresponding correlations between the estimates and references provided by PALA in (a) the kidney dataset and (b) the tumor dataset.}
    \label{fig:ulm_dv}
\end{figure*}

\subsection{In-Vivo Experiments}
Fig.~\ref{fig:roi_enlarge_rat_kidney} shows the power Doppler images of the rat kidney obtained by the tested methods. Two ROIs are zoomed in for a clearer interpretation. U\textsuperscript{2}-rPCA and IRLS-rPCA achieve better clutter suppression than SVD and 3D-Res-UNet, showing more highlighted vessels and lower tissue signals. Since 3D-Res-UNet was trained using the results of SVD, its performance is similar to the SVD filter. A closer observation on the zoomed-in regions of Fig.~\ref{fig:roi_enlarge_rat_kidney}(a) and Fig.~\ref{fig:roi_enlarge_rat_kidney}(b) demonstrates that U\textsuperscript{2}-rPCA better extracts the microvessels from surrounding tissues, showing more distinct differences between microvessels and tissues. 

\renewcommand{\arraystretch}{1.2} 
\begin{table*}[]
    \setlength\tabcolsep{12.5pt} 
    \centering
    \caption{Frame rate (fps) of the tested methods in the rat kidney dataset and the mouse tumor dataset.}
    \begin{tabular}{c|c|c|c|c|c|c}
     \toprule
    Dataset & Platform & SVD & IRLS-rPCA & 3D-Res-UNet & U\textsuperscript{2}-rPCA w/o SEU & U\textsuperscript{2}-rPCA w/ SEU  \\
        \midrule
     \multirow{2}{*}{Rat kidney}&     CPU&  83.70$\pm$19.17 & 29.95$\pm$0.47 & 1020.16$\pm$48.50 & 340.56$\pm$7.82 & 31.45$\pm$0.64\\
          \cline{2-7}
          &GPU& /& /& 2146.36$\pm$238.68 & 1307.30$\pm$417.25 & 136.61$\pm$14.37 \\
           \midrule
          \multirow{2}{*}{Mouse tumor}&     CPU&  91.88$\pm$14.94 & 137.85$\pm$7.20  & 1021.81$\pm$49.14  & 942.74$\pm$97.12 & 78.96$\pm$3.56\\
         \cline{2-7}
          &GPU& /& /& 2201.53$\pm$174.55& 2273.67$\pm$496.27 &163.84$\pm$5.50\\
         \bottomrule
    \end{tabular}
    
    \label{tab:performance}
\end{table*}

Fig.~\ref{fig:roi_enlarge_mouse_tumor} shows the power Doppler images of the mouse tumor with two zoomed-in ROIs. The results of this dataset are similar to those of the kidney dataset. The rPCA-based methods (U\textsuperscript{2}-rPCA and IRLS-rPCA) generally achieve better clutter filtering effects than the SVD-based methods (SVD and 3D-Res-UNet). Furthermore, U\textsuperscript{2}-rPCA separates the microvessels more clearly than IRLS-rPCA. 



Table.~\ref{cnr_snr} lists the CNR and SNR of the power Doppler images. Since CNR and SNR are region-dependent local metrics, we delineated two ROIs for each dataset. The delineation of ROI obeys the following principles. First, the tissue ROIs lie in the regions without visible blood flow signals. Second, the blood flow ROIs are restricted in the interior of vessels which are resolved by all the tested methods. Finally, the tissue ROIs and blood flow ROIs have similar areas. Table.~\ref{cnr_snr} clearly illustrates that U\textsuperscript{2}-rPCA achieves the highest CNR and SNR on both datasets among the tested methods. Particularly, in the kidney dataset, U\textsuperscript{2}-rPCA improves CNR by over 8.48 dB when compared with SVD and 3D-Res-UNet (ROI 2), and over 2.22 dB when compared with IRLS-rPCA (ROI 2). In the tumor dataset, the improvements of CNR achieved by U\textsuperscript{2}-rPCA are over 2.87 dB versus SVD and 3D-Res-UNet (ROI 2), and over 1.91 dB versus IRLS-rPCA (ROI 1).   

%

Fig.~\ref{fig:signal intensity} plots the cross-sectional signal intensities indicated by the white dashed lines. The profiles of the ULM density maps provided by PALA are also presented as references. One can see that U\textsuperscript{2}-rPCA and IRLS-rPCA suppress more tissue signals than SVD and 3D-Res-UNet, showing a lower level of side lobes. When inspecting the main lobes, it can be roughly justified that all the tested methods depict the vessels with comparable main-lobe widths. The profiles demonstrate that SVD-based and rPCA-based clutter filtering methods achieve similar spatial resolutions of power Doppler images. 

\begin{figure*}[!htb]
\centering
    \includegraphics[width=1\textwidth]{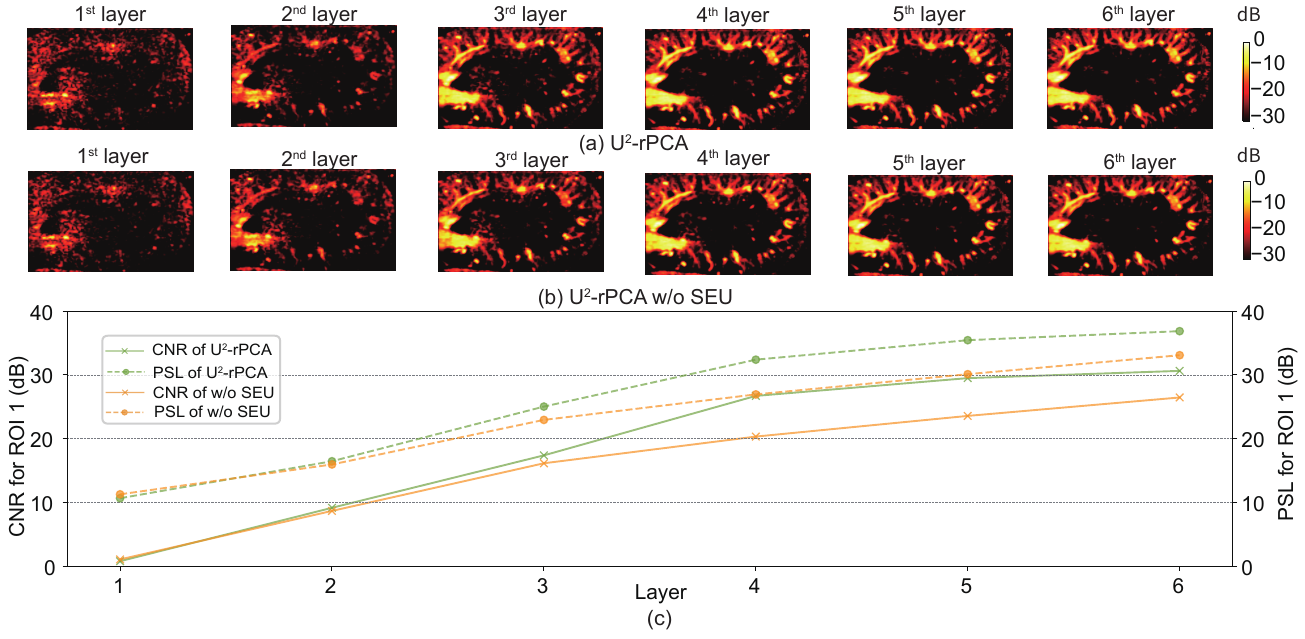}
    \caption{Influences of the number of layers $K$ on U\textsuperscript{2}-rPCA in the kidney dataset. (a) and (b) list the layer-by-layer power Doppler images output by U\textsuperscript{2}-rPCA and U\textsuperscript{2}-rPCA w/o SEU, respectively. (c) is the corresponding CNR and PSL curves.}
    \label{fig:layer_by_layer_kidney}
\end{figure*}

\begin{figure*}[!htb]
\centering
    \includegraphics[width=1\textwidth]{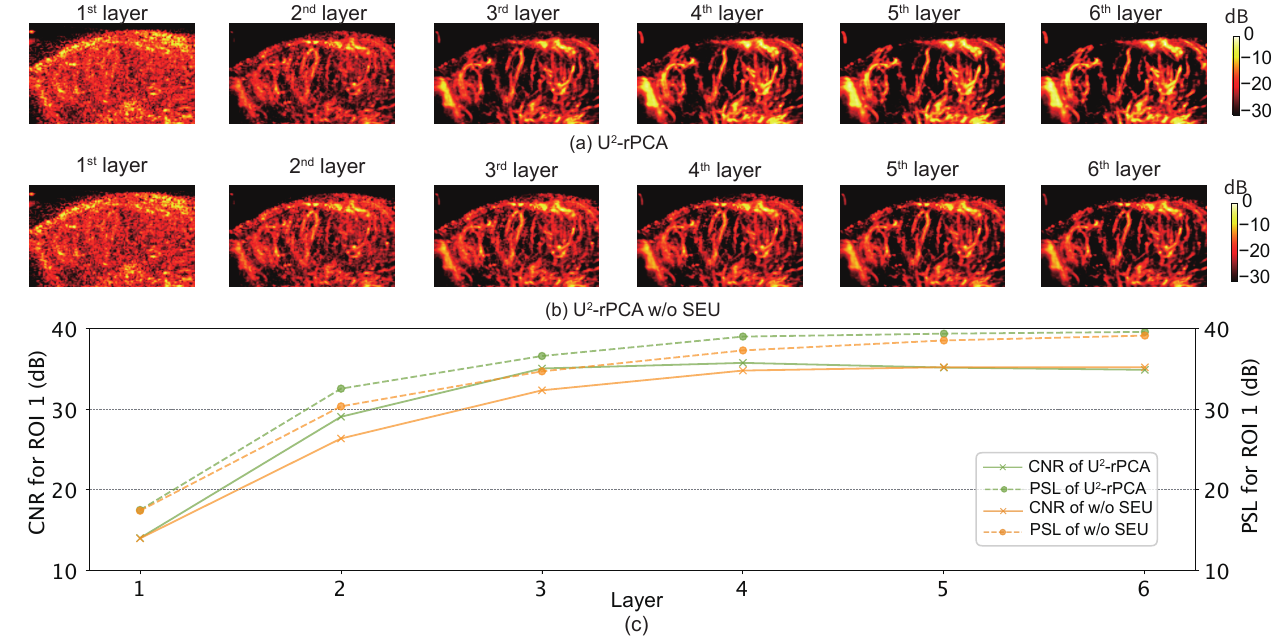}
    \caption{Influences of the number of layers $K$ on U\textsuperscript{2}-rPCA in the tumor dataset. (a) and (b) list the layer-by-layer power Doppler images output by U\textsuperscript{2}-rPCA and U\textsuperscript{2}-rPCA w/o SEU, respectively. (c) is the corresponding CNR and PSL curves.}
    \label{fig:layer_by_layer_tumor}
\end{figure*}

Fig.~\ref{fig:ulm_dv} presents the Doppler velocities obtained by SVD, IRLS-rPCA, and U\textsuperscript{2}-rPCA in the two datasets. Visual inspection and the reported correlations demonstrate that the tested methods have similar estimations. This is rational because there is no evidence to show that SVD or rPCA can provide more accurate Doppler velocities than the other. In addition, the fitted lines reveal underestimations of the Doppler velocities compared to the references. This is because the size of the microvessel is relatively small compared to the beam width in the elevation dimension~\cite{underestimation}. 

Table~\ref{tab:performance} lists the average computational time (frames per second, fps) achieved by the tested methods. The inference time of 3D-Res-UNet and U\textsuperscript{2}-rPCA was measured on CPU and GPU platforms. In summary, all methods meet the basic requirement of real-time implementation. Particularly, the SEU module has a negative impact on the inference efficiency, showing a deceleration factor of around 10 when this module is plugged into the U\textsuperscript{2}-rPCA networks.

\begin{figure}[!htb]
\centering
    \includegraphics[width=0.5\textwidth]{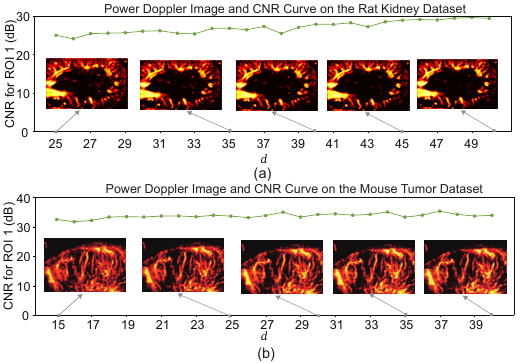}
    \caption{The influences of the inner dimension $d$ on U\textsuperscript{2}-rPCA in (a) the kidney dataset and (b) the tumor dataset.}
    \label{fig:rank_study}
\end{figure}

\subsection{Parameter Study}

Fig.~\ref{fig:layer_by_layer_kidney} and ~\ref{fig:layer_by_layer_tumor} present the layer-by-layer power Doppler images output by U\textsuperscript{2}-rPCA without and with SEU. The power Doppler images are improved when $K$ increases from 1 to 6. Particularly, in Fig.~\ref{fig:layer_by_layer_tumor}, the outputs of U\textsuperscript{2}-rPCA and the corresponding CNR and PSL curves become stable when $K$ is larger than 3. In Fig.~\ref{fig:layer_by_layer_kidney}, although the CNR and PSL curves slightly increase when $K$ is larger than 3, the visual inspection shows little variance. Therefore, we finally adopted $K=5$ for the kidney dataset and $K=4$ for the tumor dataset.
Fig.~\ref{fig:layer_by_layer_kidney} and \ref{fig:layer_by_layer_tumor} also demonstrate the effectiveness of SEU. Specifically, Fig.~\ref{fig:layer_by_layer_kidney} shows an improvement of around 5 dB in PSL with SEU when $K$ is larger than 3. In the tumor dataset, the average improvement achieved with SEU is around 2 dB when $K$ is larger than 3.

Fig.~\ref{fig:rank_study} investigates the influences of the inner dimension $d$ on U\textsuperscript{2}-rPCA. Considering that a larger $d$ allows the basis $\mathbf{U}$ to store more features about the tissue signals, we investigated the performance of U\textsuperscript{2}-rPCA by increasing $d$ from 25 to 50 in kidney and from 15 to 40 in tumor. However, Fig.~\ref{fig:rank_study} shows limited influences of this hyperparameter on U\textsuperscript{2}-rPCA, demonstrating the robustness of this method to the inner dimension. As a result, we empirically chose $d=40$ and $d=38$ for the kidney and tumor datasets, respectively.


\section{Discussion}
This paper proposes U\textsuperscript{2}-rPCA that performs clutter filtering via unsupervised unfolded deep learning. The network is unfolded from an IRLS-rPCA baseline in which both the low-rank and sparse terms are modeled with differential Frobenius norms without losing interpretability. Experimental validations on the \textit{in-silico} and public \textit{in-vivo} datasets demonstrated the performance of U\textsuperscript{2}-rPCA. In the following section, we present more insights into the effectiveness of U\textsuperscript{2}-rPCA and discuss the limitations and perspective of this paper.

%
%

   
    

           

  
    

\subsection{Effectiveness Analysis}
\begin{figure}[t]
\centering
    \includegraphics[width=0.5\textwidth]{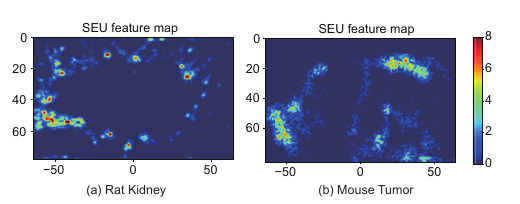}
    \caption{Visualization of feature maps extracted from SEU in (a) the kidney dataset and (b) the tumor dataset.}
    \label{fig:SEU}     
\end{figure}

Unsupervised training is one of the most representative characteristics of U\textsuperscript{2}-rPCA. Nevertheless, unsupervised training is a double-edged sword. On the one hand, the filter is insusceptible to the training labels, particularly when the \textit{in-vitro} and \textit{in-vivo} ground truths are inaccessible. On the other hand, the network behaves similarly to its rPCA baseline because it has no opportunity to learn from high-quality references. To this end, we added SEU to each layer to strengthen the network's capability in resolving the sparse microflow signals. The effectiveness of SEU is demonstrated in Figs.~\ref{fig:layer_by_layer_kidney} and \ref{fig:layer_by_layer_tumor}. Furthermore, Fig.~\ref{fig:SEU} visualizes the feature maps output by SEU in the two datasets. It can be seen that SEU focuses more on the flow regions, which is the main reason why SEU improves the quality of power Doppler images.

\begin{figure}[t]
\centering
    \includegraphics[width=0.5\textwidth]{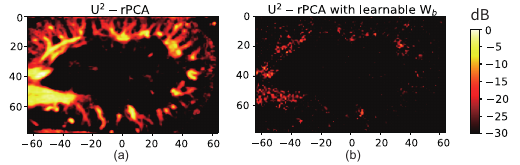}
    \caption{ Power Doppler images obtained by (a) U\textsuperscript{2}-rPCA and (b) U\textsuperscript{2}-rPCA with learnable $\mathbf{W}_b$ in the kidney dataset.}
    \label{fig:wb_learnable}
\end{figure}

However, adding SEU results in longer inference time (see Table \ref{tab:performance}). This is because the SEU module contains extensive learnable parameters. The U\textsuperscript{2}-rPCA network has 3.1M and 3.08M learnable parameters per layer with SEU for the kidney and tumor dataset, respectively. However, this number significantly decreases to 1.6k and 1.45k per layer when SEU is removed. As a result, we recommend setting SEU as a plug-in module. Whether to use this module or not depends on the task's priority, e.g., image quality or computational efficiency.



One may wonder why U\textsuperscript{2}-rPCA sets $\boldsymbol{W}_b$ as deterministic and $\boldsymbol{W}_c$ as learnable. This is because $\boldsymbol{W}_b$ is strongly connected to the blood flow signals $\boldsymbol{B}$ (see Eq.~(\ref{w_b})). The IRLS relationship between $\boldsymbol{B}$ and $\boldsymbol{W}_b$ is mathematically equivalent to the $l_1$-norm modeling. Setting $\boldsymbol{W}_b$ as deterministic parameters applies intrinsic sparse regularization to U\textsuperscript{2}-rPCA. To experimentally validate the above analyses, Fig.~\ref{fig:wb_learnable} presents the power Doppler images with deterministic and learnable $\boldsymbol{W}_b$, in which the latter setup results in the failure of U\textsuperscript{2}-rPCA. On the other hand, the low rankness of U\textsuperscript{2}-rPCA is regularized by the tissue basis $\boldsymbol{U}$ since $\boldsymbol{U}$ is itself low-rank regardless of $\boldsymbol{W}_c$. The intrinsic low-rank and sparse regularization make the network mathematically interpretable. As a result, U\textsuperscript{2}-rPCA can be trained by solely adopting the data consistency term as the loss function (see Eq.~(\ref{loss})).

\subsection{Limitation and Perspective}


Although U\textsuperscript{2}-rPCA relieves us from extensively collecting and annotating training data, the network cannot learn abundant features of the referenced tissue and blood flow signals. However, the current supervised clutter filters are limited by high-quality ground truths. The strategies of convolving moving scatterers with spatially-invariant PSF cannot fully mimic complex microvascular imaging conditions, whereas the \textit{in-vitro} and \textit{in-vivo} ground truths are technically inaccessible. Nevertheless, the potential of supervised deep filtering should be treated seriously. Recently, several realistic contrast-enhanced ultrasound simulators have been proposed to mimic the morphology and hemodynamics of microvascular networks and their interactions with ultrasound~\cite{lerendegui2022bubble,blanken2024proteus, Blanken2022SuperResolvedML, Belgharbi2023AnAR}. These simulators could help the deep learning-based methods reach a better balance in suppressing clutter and preserving blood flow signals.




\section{CONCLUSION}
This paper proposes the U\textsuperscript{2}-rPCA clutter filter for ultrasound microvascular imaging. The network was unfolded from an IRLS-rPCA baseline and trained in an unsupervised manner with mathematical interpretability. Experimental validations on the \textit{in-silico} and public \textit{in-vivo} datasets demonstrated the outperformance of U\textsuperscript{2}-rPCA when compared with the SVD method, the rPCA baseline, and another deep learning-based clutter filter. The effectiveness of the modules that construct U\textsuperscript{2}-rPCA was validated through ablation studies. Improving the training scheme and introducing semi-supervised learning to strengthen the performance of extracting micro blood flow signals are the future work of this paper. 
%
%


\bibliography{references}
\bibliographystyle{IEEEtran} 

\end{document}